\documentclass[letterpaper, 10 pt, conference]{ieeeconf}

\IEEEoverridecommandlockouts
\usepackage{graphicx}
\usepackage{times}
\usepackage{amsmath}
\usepackage{amssymb}
\usepackage{bm}
\usepackage{xcolor}

\usepackage{tikz}
\usepackage{fontawesome5}
\usetikzlibrary{fit, positioning, shapes, arrows.meta}

\usepackage{mathtools}
\mathtoolsset{showonlyrefs}

\usepackage{hyperref}

\newcommand{\Exp}[1]{\mathbb{E}\left[ #1 \right]}
\newcommand{\Cov}[1]{\mathbb{C}\mathrm{ov}\left[ #1 \right]}
\newcommand{\der}{\mathrm{d}}
\newcommand{\prob}[1]{p\left( #1 \right)}
\newcommand{\Real}[1]{\mathbb{R}^{#1}}

\newcommand{\hf}{\mathrm{hf}}
\newcommand{\lf}{\mathrm{lf}}

\newcommand\copyrighttext{%
    \footnotesize \textcopyright \the\year{} IEEE. Personal use of this material is permitted. Permission from IEEE must be obtained for all other uses, including reprinting/republishing this material for advertising or promotional purposes, collecting new collected works for resale or redistribution to servers or lists, or reuse of any copyrighted component of this work in other works.
}

\newcommand\copyrightnotice{%
    \begin{tikzpicture}[remember picture,overlay]
    \node[anchor=south,yshift=5pt] at (current page.south) {\fbox{\parbox{\dimexpr0.8\textwidth-\fboxsep-\fboxrule\relax}{\copyrighttext}}};
    \end{tikzpicture}%
}

\makeatletter
\let\NAT@parse\undefined
\makeatother

\title{\LARGE \bf
Efficient Controller Learning from Human Preferences and \\ Numerical Data via Multi-Modal Surrogate Models
}

\author{Lukas Theiner$^{1}$, Maik Pfefferkorn$^{1}$, Yongpeng Zhao$^{1,2}$, Sebastian Hirt$^{1}$ and Rolf Findeisen$^{1}$%
\thanks{$^{1}$Control and Cyber-Physical Systems Laboratory, Technical University of Darmstadt, Darmstadt, Germany, \{\mbox{rolf.findeisen}, \mbox{lukas.theiner}, \mbox{maik.pfefferkorn}, \mbox{sebastian.hirt}\}\mbox{@iat.tu-darmstadt.de}}%
\thanks{$^{2}$Volkswagen AG, Wolfsburg, Germany, \mbox{yongpeng.zhao@volkswagen.de}}%
\thanks{This work was supported by Volkswagen AG.}%
}%

\begin{document}

\maketitle
\thispagestyle{empty}
\pagestyle{empty}

\begin{abstract}
Tuning control policies manually to meet high-level objectives is often time-consuming.
Bayesian optimization provides a data-efficient framework for automating this process using numerical evaluations of an objective function. However, many systems --- particularly those involving humans --- require optimization based on subjective criteria.
Preferential Bayesian optimization addresses this by learning from pairwise comparisons instead of quantitative measurements, but relying solely on preference data can be inefficient.
We propose a multi-fidelity, multi-modal Bayesian optimization framework that integrates low-fidelity numerical data with high-fidelity human preferences.
Our approach employs Gaussian process surrogate models with both hierarchical, autoregressive and non-hierarchical, coregionalization-based structures, enabling efficient learning from mixed-modality data.
We illustrate the framework by tuning an autonomous vehicle's trajectory planner, showing that combining numerical and preference data significantly reduces the need for experiments involving the human decision maker while effectively adapting driving style to individual preferences.
\end{abstract}

\copyrightnotice

\section{Introduction}

Control policies, derived from either traditional methods or optimization-based approaches such as model predictive control (MPC), often require tuning to meet higher-level objectives.
This tuning process is often performed manually by experts, which can be time-consuming, costly, and error-prone. 
More recently, there has been a growing interest in methods that automate the learning of controller parameters to optimize closed-loop performance \cite{alonso2016_lqr_bo,berkenkamp2016_safe_controller_bo,piga2019_bo_mpc_calibration,hirt2024_stability_informed_bo,hirt2024_safe,Holzmann2024,hirt2025_hierachical}.
While various approaches have been proposed to address this challenge, Bayesian optimization (BO) \cite{garnett2023_bo} has gained particular attention, especially in settings where data collection is expensive or limited.
BO offers a data-efficient, black-box optimization framework that operates on a user-defined higher-level objective function and employs data-driven surrogate models, typically Gaussian processes (GPs) \cite{rasmussen2006}, to guide the search for optimal controller parameters.
In doing so, it can significantly reduce the number of costly closed-loop experiments required while effectively improving control performance.

In some applications, however, defining a suitable objective function for the higher-level goal is not straightforward.
This challenge arises particularly in systems that directly interact with humans, where the user’s subjective perception becomes the decisive performance criterion.
Examples include assistive devices such as exoskeletons and rehabilitation robots \cite{Ingraham2023_preference_based_in_exosceletons}, collaborative robotic systems \cite{roveda2023_huma_robot_tuning}, and automated driving applications \cite{bemporad2021_preferencebasedMPCcalibration, theiner2025_preflearnAD}.
For such cases, preferential Bayesian optimization (PBO) \cite{gonzalez2017_pbo} provides a promising alternative.
PBO extends classical Bayesian optimization by constructing its surrogate model not from quantitative evaluations of an objective function, but from qualitative user feedback in the form of pairwise preference comparisons \cite{wei2005_preference_gp,brochu2007_active_preference_learning}.
This allows optimization to be guided by subjective assessments of performance, making it particularly suitable for scenarios involving humans.

However, learning controller parameters solely from preference data can be inefficient in practice, since each preference query conveys much less information than a numerical evaluation. 
Consequently, a large number of closed-loop experiments with a human decision maker may be required, especially in high-dimensional parameter spaces.
Motivated by these challenges, we generalize and extend previous approaches by proposing a systematic framework for using multiple information sources, including those that provide preferential feedback.

Classical Bayesian optimization methods, which rely exclusively on numerical evaluations of the objective function, have been extensively studied in the context of multiple information sources \cite{kennedy2000_ar1,poloczek2017_multiple_information_source,forrester2007_multifidelity_bo,swersky2013_multitask}.
A common category of these methods is multi-fidelity approaches, in which one or more lower-fidelity sources are available alongside the main objective function. Such sources are typically cheaper to evaluate but provide only an approximate assessment of the objective.
These settings arise frequently, for example, when a costly real-world experiment can be supplemented by a cheaper simulation that does not perfectly match real-world conditions.
For instance, a hierarchical GP surrogate model structure was employed in \cite{yongpeng2025_multifidelitybo,zhao2026_constrained_mfbo} to combine simulation and real-world data for vehicle controller tuning.
Similarly, policy tuning for robotic systems using both simulation and experimental data has been investigated in \cite{He2025}, where, however, the GP surrogate model does not impose a hierarchical relationship between the sources.
Several other multi-fidelity Bayesian optimization methods have been adapted to address settings where auxiliary information sources exhibit varying or poor quality \cite{Nobar2026_guided_mfbo_digital_twin,fan2024_mfbo_input_dependent,mikkola2023_mfbo_unreliable}.

In this work, we build on established multi-fidelity approaches by considering hierarchical and non-hierarchical surrogate model structures that incorporate multiple information sources.
Our framework extends classical multi-fidelity optimization to scenarios where the available sources differ not only in fidelity but also in modality, including those that provide preferential feedback rather than numerical evaluations.
Specifically, we develop multi-modal GP models to integrate numerical and preferential data. 
These models are based on both hierarchical, autoregressive structures and non-hierarchical, coregionalization-based structures. 
We discuss computational aspects related to model training and inference and describe their application within BO to optimize higher-level objectives using both numerical and preferential data.
Finally, we demonstrate our proposed approach in the context of tuning a trajectory planner for autonomous driving. 
In this setting, numerical data act as low-fidelity information, derived from a fixed-structure, approximate performance measure, while preferential data serve as high-fidelity information, obtained directly from human drivers or passengers.

The remainder of this article is organized as follows. 
In Section \ref{sec:problem_formulation}, we formalize the considered setting. 
Section \ref{sec:gpr_bo} introduces Gaussian process models based on numerical or preferential data and provides an overview of classical multi-fidelity model structures, along with a brief introduction to Bayesian optimization. 
In Section \ref{sec:num_pref_gpr}, we propose multi-fidelity GP models that integrate both numerical and preferential data for use in PBO.
Section \ref{sec:example} presents an application of the proposed models to trajectory planning for autonomous driving within the PBO framework. Finally, we conclude in Section \ref{sec:conclusions}.

\section{Problem formulation}\label{sec:problem_formulation}

We consider a controlled dynamical system 
\begin{equation} \label{eq:system}
    x_{t+1} = f\big(x_t, u_t), \; u_t = \pi_\xi(x_t)
\end{equation}
with state $x_t \in \mathcal{X} \subset \Real{n_\mathrm{x}}$, input $u_t \in \mathcal{U} \subset \Real{n_\mathrm{u}}$, and time index $t \in \mathbb{N}$. 
The system dynamics are described by a function $f: \mathcal{X} \times \mathcal{U} \rightarrow \mathcal{X}$, and the system is controlled by a parameterized policy $\pi_\xi: \mathcal{X} \rightarrow \mathcal{U}$, where $\xi \in \mathcal{I} \subset \Real{n_\xi}$ denotes the adjustable parameters.
Applying the control policy $\pi_\xi$ in the closed loop for $T \in \mathbb{N}$ steps from an initial state $x_0$ generates a closed-loop trajectory $\tau(\xi, x_0)=(\{x_t\}_{t\in [0,\dots,T]},\{u_t\}_{t\in [0,\dots,T-1]})$.

The dynamical system interacts with a human decision maker, who evaluates the performance of the closed-loop trajectory $\tau(\xi, x_0)$ for a given parameter set $\xi$ according to a latent, unknown objective function 
\begin{equation} \label{eq:true-objective-function}
    G: \Real{n_\xi} \rightarrow \Real{}, \xi \mapsto G(\xi).
\end{equation}
The goal is to identify parameters $\xi$ that optimize this latent objective function, which reflects the decision maker’s preferences. To this end, the decision maker can be queried to express pairwise preferences between two parameter configurations $(\xi_a, \xi_b)$ after observing the corresponding system behaviors.
Since each query requires a physical experiment involving the human, the total number of queries is limited by the experimental budget, making data-efficient learning of the optimal parameters essential.
To improve data efficiency, an additional information source 
\begin{equation} \label{eq:approx-objective-functions}
    \hat{G}: \Real{n_\xi} \rightarrow \Real{}, \xi \mapsto \hat{G}(\xi)
\end{equation}
is assumed to be available.
This may include, for example, \pagebreak a simulation model of the controlled dynamical system \eqref{eq:system} and an approximation of the decision maker’s evaluation process \eqref{eq:true-objective-function}.
While system models are commonly available in control design, modeling the decision maker depends on the application context and may reflect characteristic human behavior or factors such as comfort, cognitive effort, or perceived safety, based on simulated or measured quantities.
To improve data efficiency, the outputs of the additional information source \eqref{eq:approx-objective-functions} are assumed to be correlated with the original objective function \eqref{eq:true-objective-function}.

\begin{figure}[t]
    \centering
    \vspace{2mm}
    \newcommand{\myfigsize}{\small}

\begin{tikzpicture}[
    block/.style={rectangle, draw, minimum width=1.4cm, minimum height=0.6cm, align=center},
    colorblock/.style={block, minimum width=0.8cm, minimum height=0.6cm, 
        draw=black,
        fill=black,
        fill opacity=0.1,
        text opacity=1,
        rounded corners,
    },
    redblock/.style={colorblock, draw=red!50!black, fill=red!80!black},
    greenblock/.style={colorblock, draw=green!50!black, fill=green!80!black},
    line/.style={-{Latex}},
    dashedarrow/.style={-latex, double, dashed}
]

\node[block] (system1) {\myfigsize$f(x_t, u_t)$};
\node[block, below=0.2cm of system1] (policy1) {\myfigsize$\pi_\xi(x_t)$};
\draw[line] (system1.east) -- ++(0.3,0) |- node[pos=0.25, right] {\myfigsize$x_t$} (policy1.east);
\draw[line] (policy1.west) -- ++(-0.3,0) |- node[pos=0.25, left] {\myfigsize$u_t$} (system1.west);

\coordinate (clsys1_topright) at ([xshift=0.7cm,yshift=0.3cm]system1.east);
\coordinate (clsys1_bottomleft) at ([xshift=-0.7cm,yshift=-0.3cm]policy1.west);
\node[
    greenblock,
    fit=(clsys1_bottomleft)(clsys1_topright),
    label={[green!50!black]below:{\myfigsize High-fidelity Experiment}}
] (highbox) {};

\node[block, right=2cm of system1] (system2) {\myfigsize$f(x_t, u_t)$};
\node[block, below=0.2cm of system2] (policy2) {\myfigsize$\pi_\xi(x_t)$};
\draw[line] (system2.east) -- ++(0.3,0) |- node[pos=0.25, right] {\myfigsize$x_t$} (policy2.east);
\draw[line] (policy2.west) -- ++(-0.3,0) |- node[pos=0.25, left] {\myfigsize$u_t$} (system2.west);

\coordinate (clsys2_topright) at ([xshift=0.7cm,yshift=0.3cm]system2.east);
\coordinate (clsys2_bottomleft) at ([xshift=-0.7cm,yshift=-0.3cm]policy2.west);
\node[
    redblock,
    fit=(clsys2_bottomleft)(clsys2_topright),
    label={[red!50!black]below:{\myfigsize Low-fidelity Simulation}}
] (lowbox) {};

\coordinate (cltraj1) at ([xshift=0.5cm]highbox.north);
\coordinate (cltraj2) at ([xshift=0.5cm]lowbox.north);
\coordinate (xi1b) at ([xshift=-0.5cm]highbox.north);
\coordinate (xi2b) at ([xshift=-0.5cm]lowbox.north);

\node[greenblock, above=0.7cm of cltraj1] (G1) {\myfigsize \faUser};
\node[redblock, above=0.7cm of cltraj2] (G2) {\myfigsize$\hat{G}$};

\draw[dashedarrow] (cltraj1) -- node[pos=0.5, right] {\myfigsize$\tau(\xi, x_0)$} (G1.south);
\draw[dashedarrow] (cltraj2) -- node[pos=0.5, right] {\myfigsize$\hat\tau(\xi, x_0)$} (G2.south);

\coordinate (mfbo_right) at ([xshift=0.7cm]system2.east);
\coordinate (mfbo_left) at ([xshift=-0.7cm]system1.west);
\node[colorblock, 
    fit=(mfbo_right)(mfbo_left),
    minimum height=1cm,
    yshift=2.9cm,
   ] (MMMFBO) {\myfigsize Multi-Modal Multi-Fidelity \\ Bayesian Optimization};

\draw[dashedarrow] (MMMFBO.south -| xi1b) -- node[pos=0.5, left] {\myfigsize$\xi_a, \xi_b$} (xi1b);
\draw[dashedarrow] (MMMFBO.south -| xi2b) -- node[pos=0.5, left] {\myfigsize$\xi$} (xi2b);

\draw[dashedarrow] (G1.north) -- node[pos=0.5, right] {\footnotesize \faThumbsUp\;\faThumbsDown} (MMMFBO.south -| G1.north);
\draw[dashedarrow] (G2.north) -- node[pos=0.5, right] {\myfigsize$\hat G(\xi)$} (MMMFBO.south -| G2.north);

\end{tikzpicture}
    \caption{Structure of the proposed multi-modal multi-fidelity BO framework. Due to the human decision maker, the high-fidelity experiment only returns preferential data. The additional information source provides low-fidelity numerical evaluations.}
    \label{fig:overview_general}
    \vspace{-3mm}
\end{figure}
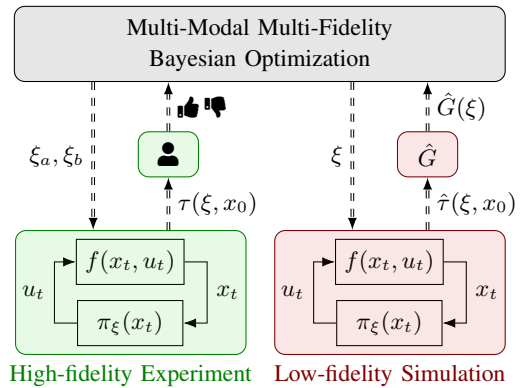

The approach presented herein generalizes naturally to multiple additional information sources; for simplicity of presentation, we restrict the analysis to one such source.
The considered setting is visualized in Figure \ref{fig:overview_general}.

\section{Gaussian Processes \& Bayesian Optimization} \label{sec:gpr_bo}

To optimize the high-level objective, we employ Bayesian optimization, which utilizes a surrogate model of the objective function --- commonly a Gaussian process.
This section therefore outlines the fundamentals of Gaussian process modeling, covering both the standard regression setting and the case where a latent function is inferred solely from preferential observations.
In addition, we summarize two established approaches for constructing Gaussian process models that integrate multiple numerical information sources, and conclude with a brief overview of Bayesian optimization.

\subsection{Basics of Gaussian Process Regression} \label{sec:fundamentals_gp}

Gaussian processes generalize the concept of Gaussian random vectors to infinite-dimensional function spaces and are commonly interpreted as Gaussian probability distributions over functions.
Formally, a GP is denoted by 
\begin{align}\label{eq:gp_prior}
    g(\xi) \sim \mathcal{GP}\left( m_\theta(\xi), k_\theta(\xi, \xi') \right),
\end{align}
and defines a collection of random variables $\{ g(\xi) \mid \xi \in \mathcal{I} \subseteq \Real{n_\mathrm{\xi}} \}$, indexed by $\xi$, any finite number of which are jointly Gaussian distributed \cite{rasmussen2006}.
A Gaussian process model is fully defined by the prior mean function $m_\theta: \Real{n_\mathrm\xi} \rightarrow \Real{}, \xi \mapsto \Exp{g(\xi)}$ and the prior covariance function $k_\theta: \Real{n_\mathrm\xi} \times \Real{n_\mathrm\xi} \rightarrow \Real{}, (\xi, \xi^\prime)\mapsto \Cov{g(\xi),g(\xi^\prime)}$, which both depend on a set of hyperparameters $\theta$.
\newline
Given fixed hyperparameters $\theta$, evaluating \eqref{eq:gp_prior} at a finite set of inputs $(\xi_1, \ldots, \xi_d)$ yields a multivariate Gaussian distribution $(g(\xi_1), \ldots, g(\xi_d)) \sim \mathcal{N}(\mu, \Sigma)$ with mean vector $\mu$, with entries $[\mu]_i = m_\theta(\xi_i)$, and covariance matrix $\Sigma$, with entries $[\Sigma]_{ij} = k_\theta(\xi_i, \xi_j)$.

In the context of Bayesian optimization, Gaussian processes are commonly used as surrogate models of the unknown objective function.
More generally, Gaussian processes are used to probabilistically model an unknown function $G(\xi): \Real{n_\mathrm\xi} \rightarrow \Real{}$ based on a finite set of prior observations.
Specifically, we rely on a data set $\mathcal{D} = (\Xi, \mathcal{O})$, where $\Xi \in \Real{n_\mathrm{\Xi} \times n_\mathrm{\xi}}$ is the input matrix with rows $[\Xi]_{i:} = \xi_i^\top$ and $\mathcal{O}$ denotes corresponding observations related to the function.
Notably, the nature of $\mathcal{O}$ depends on whether the Gaussian process model $g$ of $G$ is constructed from numerical or preferential data; this distinction will be elaborated upon shortly.
The objective is to leverage the available data together with the prior model introduced in \eqref{eq:gp_prior} to make informed predictions of the unobserved function values $\mathbf{g}_*=[G(\xi_{*,1}),\dots,G(\xi_{*,n_*})]^\top$ at test locations $\Xi_{*},\; [\Xi_*]_{i:}=\xi_{*,i}^\top$, which are jointly distributed with the latent function values $\mathbf{g}=[G(\xi_1), \dots, G(\xi_{n_\Xi})]^\top$ at training inputs $\Xi$.

To this end, given a data set $\mathcal{D}$ of observations of the underlying function $G$, we aim to infer the data-informed posterior distribution $\prob{\mathbf{g}_*  \mid  \mathcal{D}, \Xi_*}$.
To achieve this, we consider the conditional Gaussian distribution
\begin{align}\label{eq:gp_conditional}
    & \mathbf{g}_* \mid \Xi,\mathbf{g}, \Xi_*, \theta \sim \mathcal{N}(\mu_{\mathbf{g}_* \mid \mathbf{g}}, \Sigma_{\mathbf{g}_* \mid \mathbf{g}}), \quad\text{where} 
    \\
    &\mu_{\mathbf{g}_*\mid \mathbf{g}} = m_{\theta}(\Xi_*) + k_{\theta}(\Xi_*, \Xi) k_{\theta}(\Xi, \Xi)^{-1} (\mathbf{g} - m_{\theta}(\Xi)), \notag
    \\
    &\Sigma_{\mathbf{g}_*\mid \mathbf{g}} = \! k_{\theta}(\Xi_*, \Xi_*) \! - \! k_{\theta}(\Xi_*, \Xi) k_{\theta}(\Xi, \Xi)^{-1} k_{\theta}(\Xi, \Xi_*). \notag
\end{align}
Here, $m_{\theta}(A)$ denotes the vector with entries $[m_{\theta}]_i = m_{\theta}([A]_{i:})$, and $k_{\theta}(A,B)$ denotes the covariance matrix with entries $[k_{\theta}(A,B)]_{ij} = k_{\theta}([A]_{i:}^\top, [B]_{j:}^\top)$ for $A,B \in \{\Xi, \Xi_*\}$.

We further consider the likelihood $\prob{\mathcal{O}  \mid  \mathbf{g}, \theta}$, which quantifies the probability of observations $\mathcal{O}$ given the latent function values $\mathbf{g}$ under a Gaussian process model with hyperparameters $\theta$.
In general, the hyperparameters may also be treated as uncertain, with a prior distribution $\theta \sim \prob{\theta}$.
Applying Bayes' theorem, the posterior model is given by
\begin{multline}\label{eq:fully_bayesian_posterior}
    \prob{\mathbf{g}_* \mid \mathcal{D}, \Xi_*}  \propto \iint
    \prob{\mathbf{g}_* \mid \Xi,\mathbf{g}, \Xi_*, \theta}
    \,
    \prob{\mathcal{O}  \mid  \mathbf{g}, \theta}
    \\
    \prob{\mathbf{g}  \mid  \Xi,\theta}
    \,
    \prob{\theta} 
    \,
    \mathrm{d} \mathbf{g}
    \,
    \mathrm{d} \theta.
\end{multline}
The mean of the posterior distribution \eqref{eq:fully_bayesian_posterior} provides an estimate of the unknown function values $G(\xi_{*,1}), \dots,  G(\xi_{*,n_*})$, while its (co-)variance quantifies the associated prediction uncertainty.

\vspace{2mm}
\subsubsection{Numerical observations} \label{sec:fundamentals_gp_num}

In many settings, Gaussian process models are trained on numerical data.
In such cases, the observations $\mathcal{O}$ are the target vector $\mathbf{y}$ consisting of noisy evaluations of the underlying function, i.e., $[\mathbf{y}]_i = G(\xi_i) + \varepsilon_i$, where the noise terms $\varepsilon_i$ are typically assumed to be independent and identically distributed Gaussian random variables, $\varepsilon_i \sim \mathcal{N}(0, \sigma_\mathrm{n}^2)$, with zero mean and variance $\sigma_\mathrm{n}^2$.
This assumption implies a Gaussian likelihood $\prob{\mathcal{O}  \mid  \mathbf{g}, \theta}$.

Moreover, in most applications, a point estimate $\theta^*$, obtained, e.g., via evidence maximization or cross-validation, is typically used \cite{rasmussen2006}. 
Combined with the Gaussian likelihood assumption, this leads to a posterior $\prob{\mathbf{g}_*  \mid  \mathcal{D}, \Xi_*, \theta^*}$ that is itself Gaussian and can be expressed
in closed-form
\begin{align}
    & \mathbf{g}_* \mid \mathcal{D}, \Xi_*, \theta^* \sim \mathcal{N}(m^+_{\theta^*}(\Xi_*), k^+_{\theta^*}(\Xi_*, \Xi_*)),~~ \text{where} \notag \\
    & m^+_{\theta^*}(\Xi_*) = m_{\theta^*}(\Xi_*) + k_{\theta^*}(\Xi_*, \Xi) \Sigma_\mathbf{y}^{-1} (\mathbf{y} - m_{\theta^*}(\Xi)), \label{eq:gaussian_posterior_closed_form} \\
    & k^+_{\theta^*}(\Xi_*, \Xi_*) \! = \! k_{\theta^*}(\Xi_*, \Xi_*) \! - \! k_{\theta^*}(\Xi_*, \Xi) \Sigma_\mathbf{y}^{-1} k_{\theta^*}(\Xi, \Xi_*), \notag
\end{align}
with $\Sigma_\mathbf{y} = k_{\theta^*}(\Xi, \Xi) + \sigma_\mathrm{n}^2 \mathbb{I}$ denoting the covariance matrix of the observations $\mathbf{y}$.

In settings with non-Gaussian noise or uncertain hyperparameters $\theta \sim \prob{\theta}$, approximations are typically required to obtain a closed-form posterior \cite{rasmussen2006}, or alternatively, the posterior can be estimated directly using Markov-chain Monte-Carlo (MCMC) methods to evaluate either the fully Bayesian posterior $\prob{\mathbf{g}_*  \mid  \mathcal{D}, \Xi_*}$ in \eqref{eq:fully_bayesian_posterior} or the conditional posterior $\prob{\mathbf{g}_*  \mid  \mathcal{D}, \Xi_*, \theta^*}$, depending on whether a fixed or marginal treatment of hyperparameters is adopted.

\vspace{2mm}
\subsubsection{Preference observations} \label{sec:fundamentals_gp_pref}
In preference learning with Gaussian processes \cite{wei2005_preference_gp}, the latent function is modeled based on preference observations in the form of pairwise comparisons. 
Instead of noisy evaluations of the underlying function, the observations $\mathcal{O}$ are in the form of a set of $n_\mathcal{C}$ comparisons $\mathcal{C} = \{ (i_k, j_k) \mid k \in \{1, \ldots, n_\mathcal{C} \} \}$, where $i_k,j_k \in \{1,\dots,n_\Xi\}$ and $i_k \neq j_k$. Within the set $\mathcal{C}$, each tuple $(i_k, j_k)$ represents one comparative judgment by the decision maker.
In particular, the $k^\text{th}$ tuple $(i_k,j_k)$ means that the decision maker preferred $\xi_{i_k}$ over $\xi_{j_k}$.
A common modeling assumption is that comparisons are based on the value of a latent objective function, contaminated with Gaussian noise.
This implies the so-called \emph{probit} likelihood function
\begin{multline}
        \prob{(i,j)  \mid  g(\xi_i), g(\xi_j)} = \iint \mathbf{1}_{\{ g(\xi_i) + \varepsilon_i \geq g(\xi_j) + \varepsilon_j\}} (\varepsilon_i, \varepsilon_j) \, 
        \\
        p(\varepsilon_i) \, p(\varepsilon_j) \, \der \varepsilon_i \, \der \varepsilon_j ,
\end{multline}
where $\mathbf{1}_A$ denotes the indicator function for the set $A$ and $\varepsilon_i, \varepsilon_j$ are independent and identically distributed random variables $\varepsilon_i, \varepsilon_j \sim \mathcal{N}(0, \sigma_\mathrm{n}^2)$. 
Hence, the distribution $\prob{(i,j)  \mid  g(\xi_i), g(\xi_j)}$ admits the closed-form expression
\begin{equation} \label{eq:probit}
    \prob{(i,j)  \mid  g(\xi_i), g(\xi_j)} = \Phi\left(\frac{g(\xi_i) - g(\xi_j)}{\sqrt{2\sigma_\mathrm{n}^2}}\right),
\end{equation}
where $\Phi$ is the cumulative distribution function of the standard normal distribution.
Notably, unlike with numerical observations, the posterior distribution for preferential data is not Gaussian, even if the hyperparameters are known.
It can either be approximated as Gaussian using the Laplace approximation \cite{wei2005_preference_gp, rasmussen2006} or estimated via sampling-based approaches, such as MCMC methods.

To obtain a sampling-based approximation of the posterior \eqref{eq:fully_bayesian_posterior}, we first generate $S$ posterior samples of the latent function values at training locations and hyperparameters, $(\mathbf{g}^{(s)}, \theta^{(s)}) \sim p(\mathbf{g}, \theta \mid \mathcal{D})$, for $s=1,\dots,S$. \newline
Next, for each sample, we compute the conditional distribution of the test point, $\mathbf{g}_* \mid \Xi,\mathbf{g}^{(s)}, \xi_*, \theta^{(s)}$, as defined in \eqref{eq:gp_conditional}.
Finally, resampling from these conditional distributions yields posterior predictive samples of $\mathbf{g}_*$.

\subsection{Multi-Fidelity Gaussian Processes}

Traditional Gaussian process regression, as discussed earlier, is designed to model data from a single, typically high-fidelity, source and assumes that all observations are of consistent quality.
However, in many practical applications, acquiring high-fidelity data, such as through physical experiments or real-world testing, can be costly, time-intensive, or constrained by logistical limitations.

To mitigate the reliance on expensive data, low-fidelity simulations are often employed as a complementary, less resource-intensive alternative.
This introduces the challenge of integrating heterogeneous data sources with differing levels of fidelity, which has led to the development of extended GP regression frameworks for multi-fidelity settings \cite{Brevault2020}.

In the following, we provide a concise summary of two established strategies to model multiple data sources $G_h(\xi), \;h \in \{ 1,\ldots,H \}$ through a collection of GPs $g_h(\xi), \; h \in \{ 1,\dots,H \}$. 
Importantly, these methods impose a correlation between the GP models, allowing for information transfer between them.  

\vspace{2mm}
\subsubsection{Intrinsic Coregionalization Model (ICM)} \label{sec:fundamentals_icm}
A common approach to constructing surrogate models that integrate multiple information sources is the use of \emph{coregionalization} \cite{bonilla2007_multitask,swersky2013_multitask}, as also employed in multi-output Gaussian process regression \cite{alvarez2012_kernel_multioutput}.
In this framework, each information source is modeled as a separate but correlated Gaussian process $g_h$, where dependencies are induced through a shared set of $H$ latent Gaussian processes,
\begin{equation}
    u_i(\xi) \sim \mathcal{GP}\bigl(m(\xi), k(\xi,\xi')\bigr), \quad i \in \{1, \dots, H\},
\end{equation}
that all share a common kernel function $k$ but are mutually independent.
Each model $g_h$ is a linear combination of the latent processes,
\begin{equation}
    g_h(\xi) = \sum_{i=1}^{H} a_{h,i} \, u_i(\xi), \quad h \in \{ 1, \dots, H \},
\end{equation}
with coefficients $a_{h,i} \in \mathbb{R}$ that determine how strongly output $h$ depends on latent process $i$.  
Using the mutual independence of the latent GPs, the prior cross-covariance between $g_h$ and $g_{h'}$ can be derived as
\begin{align}
    \Cov{g_h(\xi), g_{h'}(\xi')}
    &= \sum_{i=1}^{H} a_{h,i} a_{h',i} \, k(\xi, \xi') \\
    &= [B]_{hh'} \, k(\xi, \xi'),
\end{align}
where the \emph{coregionalization matrix} $B = A A^\top$ has entries $[A]_{hi} = a_{h,i}$.
This symmetric positive semi-definite matrix captures the inter-output correlations induced by the latent mixing structure, and its entries are inferred jointly with the kernel hyperparameters \cite{swersky2013_multitask}.

The ICM can equivalently be viewed as a standard Gaussian process defined over an augmented input space $\tilde{\mathcal{I}} = \mathcal{I} \times \{1, \dots, H\}$,
with the corresponding augmented kernel
\begin{equation} \label{eq:icm_augmented_kernel}
    \tilde{k}\big( (\xi, h), (\xi', h') \big) = [B]_{hh'} \, k(\xi, \xi').
\end{equation}
Inference in this formulation proceeds analogously to the single-output GP described in Section~\ref{sec:fundamentals_gp_num}.

\vspace{2mm}
\subsubsection{Autoregressive Model (AR1)} \label{sec:fundamentals_ar1}
Secondly, we also utilize the autoregressive (AR1) model introduced in \cite{kennedy2000_ar1}, which establishes a Gaussian process prior for each fidelity level and models their interdependencies through a recursive, linear structure.

Let $G_h(\xi),\; h=1,\ldots,H$ represent a sequence of models of $G$ with increasing fidelity.
Each model's output is captured by a separate GP $g_h$, defined analogously to \eqref{eq:gp_prior}.

Specifically, at the base fidelity level $h=1$ (lowest fidelity), we define a standard Gaussian process model
\begin{align}
    g_1(\xi) \sim \mathcal{GP}\bigl(m_1(\xi), k_1(\xi,\xi')\bigr).
\end{align}
For subsequent fidelity levels $h=2, \ldots, H$, the model is recursively defined via
\begin{align}\label{eq:ar1_gp}
    \left\{
    \begin{array}{l}
        g_h(\xi)={\rho_{h-1}}g_{h-1}(\xi)+\delta_h(\xi), \\
        g_{h-1}(\xi)	\perp \delta_h(\xi), \\
    \end{array}
    \right.
\end{align}
where the correction term $\delta_h$ is itself modeled as an independent ($\perp$) Gaussian process
$\delta_h(\xi) \sim \mathcal{GP}\bigl(m_{\delta_h}(\xi), k_{\delta_h}(\xi,\xi')\bigr)$.
In this formulation, the parameters $\rho_{h-1}$ govern the linear dependence between successive fidelity levels.
Due to the closure of Gaussian processes under linear transformations, the additive form in \eqref{eq:ar1_gp} ensures that all higher fidelity models $g_h$, $h=2, \ldots, H$, are themselves Gaussian processes.

As in the single-fidelity case (Section \ref{sec:fundamentals_gp_num}), closed-form posteriors are possible under Gaussian noise and fixed hyperparameters, whereas non-Gaussian noise or uncertain hyperparameters necessitate approximation or MCMC-based Bayesian inference.

\subsection{Bayesian Optimization}

We apply Bayesian optimization to address the optimization problem
\begin{subequations}
\begin{align}
    \xi^* = \arg \min_{\xi \in \mathcal{I}} G(\xi),
\end{align}
\end{subequations}
where $G: \mathbb{R}^{n_\mathrm{\xi}} \to \mathbb{R}, \xi \mapsto G(\xi)$ is a black-box function and $\xi^*$ denotes its global minimizer within the feasible set $\mathcal{I}$.
Unlike classical optimization methods that require dense evaluations of the objective, BO seeks to minimize the number of function evaluations by sequentially selecting the most informative sampling points.
This is particularly advantageous when the true objective $G$ is unknown and expensive to evaluate.
To this end, BO relies on surrogate models of the objective $G$ --- most commonly Gaussian processes --- which are updated iteratively as new observations are gathered.
Specifically, in each iteration $n \in \mathbb{N}$, BO proceeds in two main steps:
\pagebreak
\begin{enumerate}
    \item[1)]
    Select a query point $\xi_n$ and evaluate the objective function --- which in the controller tuning case involves running a closed-loop experiment --- to obtain a new data tuple $(\xi_n, G(\xi_n))$, and
    \item[2)]
    update the GP surrogate, including hyperparameters, using the augmented data set $\mathcal{D}_{n+1} = (\Xi_{n+1}, \mathbf{y}_{n+1})$, where $\Xi_{n+1} \leftarrow [\Xi_n^\top, \xi_n]^\top$ and $\mathbf{y}_{n+1} \leftarrow [\mathbf{y}_n^\top, G(\xi_n)]^\top$.
\end{enumerate}
For Preferential Bayesian optimization (PBO), with observations in the form of pairwise comparisons, the steps are:
\begin{enumerate}
    \item[1)] 
    Select a pair of query points $(\xi_{i_n}, \xi_{j_n})$ and query the decision maker --- by running two closed-loop experiments --- to obtain a new data tuple $(\xi_{i_n}, \xi_{j_n}, c_n)$, where $c_n = (i_n, j_n)$ if $\xi_{i_n}$ is preferred over $\xi_{j_n}$ and $c_n = (j_n, i_n)$ otherwise, and
    \item[2)]
    update the GP surrogate, including hyperparameters, using the augmented data set $\mathcal{D}_{n+1} = (\Xi_{n+1}, \mathcal{C}_{n+1})$, where $\Xi_{n+1} \leftarrow [\Xi_n^\top, \xi_{i_n}, \xi_{j_n}]^\top$ and $\mathcal{C}_{n+1} \leftarrow \mathcal{C}_n \cup \{ c_n \}$.
\end{enumerate}

To steer the selection of the query points $\xi_n$ toward the global minimizer $\xi^*$, BO relies on an acquisition function $\alpha_\mathrm{BO}$.
This function uses the current GP surrogate for $G$ to quantify the utility of evaluating a candidate point, effectively balancing exploration of uncertain regions with exploitation of promising areas.
At each iteration $n$, the next point to evaluate is selected by maximizing this acquisition function:
\begin{align}
    \xi_n = \arg \max_{\xi \in \mathcal{I}} \alpha_\mathrm{BO}(\xi; \mathcal{D}_n),
\end{align}
where $\alpha_\mathrm{BO}: \mathbb{R}^{n_\mathrm{\xi}} \to \mathbb{R}, \xi \mapsto \alpha_\mathrm{BO}(\xi; \mathcal{D}_n)$.
For an overview of commonly used acquisition functions in the standard setting with numerical observations, we refer the reader to \cite{garnett2023_bo}.

In preferential Bayesian optimization, a pair of points for the next comparison is selected instead:
\begin{align}
   (\xi_{i_n}, \xi_{j_n}) = \arg \max_{(\xi, \xi') \in \mathcal{I}^2} \alpha_\mathrm{PBO}(\xi, \xi'; \mathcal{D}_n),
\end{align}
where $\alpha_\mathrm{PBO}: \mathbb{R}^{n_\mathrm{\xi}} \times \mathbb{R}^{n_\mathrm{\xi}} \to \mathbb{R}, (\xi, \xi') \mapsto \alpha_\mathrm{PBO}(\xi, \xi'; \mathcal{D}_n)$.
A widely used pairwise acquisition function is the \emph{expected utility of the best option} (EUBO) \cite{lin2022_eubo, astudillo2023_qEUBO}. For a summary of alternative acquisition functions in the preferential setting, see \cite{astudillo2023_qEUBO} and the references therein. 

\subsection{Multi-Fidelity Bayesian Optimization}\label{sec:mfbo}
In multi-fidelity Bayesian optimization, a surrogate model incorporating data from all information sources is used to improve data efficiency with respect to the main, high-fidelity source. 
Depending on the application context, two distinct strategies can be employed. 
The first strategy employs cost-aware acquisition functions to dynamically select the most informative and cost-effective fidelity at each iteration \cite{poloczek2017_multiple_information_source, swersky2013_multitask, Nobar2026_guided_mfbo_digital_twin}.
The second strategy organizes the optimization into distinct phases, starting with exploration on low-fidelity data and transitioning to high-fidelity evaluations in later stages \cite{yongpeng2025_multifidelitybo, zhao2026_constrained_mfbo}.
In this phased approach, exploration-oriented acquisition functions --- such as the \emph{integral predictive variance} (IPV) \cite{seo2000_ipv} --- should be used during the low-fidelity phase, as demonstrated by \cite{yongpeng2025_multifidelitybo}.

\section{Multi-Modal Multi-Fidelity Surrogates Combining Preferential and Numerical Data}
\label{sec:num_pref_gpr}

Multi-fidelity methods for Bayesian optimization aim to reduce the reliance on expensive data by utilizing additional lower-fidelity data sources. 
Prior work was limited to combining data sources providing numerical evaluations of the objective function.  
In this section, we extend these approaches by integrating both numerical and preferential observations of different fidelities into a unified multi-modal, multi-fidelity surrogate model.
For clarity of exposition, the following discussion considers two fidelity levels ($H=2$), without loss of generality.
Specifically, the high-fidelity level ($h=2$) corresponds to pairwise comparisons of \eqref{eq:true-objective-function} and is denoted by superscript $\hf$.
The low-fidelity level ($h=1$) corresponds to numerical observations of \eqref{eq:approx-objective-functions} and is denoted by superscript $\lf$.
As discussed earlier, this setting captures scenarios in which controller or system parameters are learned according to human preferences, while an available system model and approximate, numerically tractable decision criterion are leveraged to minimize the need for human-involved experiments.

\subsection{Multi-Modal ICM Surrogates} \label{sec:icm_mm}

We begin by discussing the implications of incorporating multi-modal data for the application of the ICM (cf. Section \ref{sec:fundamentals_icm}) as a multi-fidelity surrogate.
For clarity of exposition, we treat the ICM as a standard GP $\tilde{g}$ defined over an augmented input space and equipped with an augmented kernel $\tilde{k}$, as in \eqref{eq:icm_augmented_kernel}.
The problem thus reduces to conditioning a GP simultaneously on numerical and preference observations. 

We denote the augmented high-fidelity and low-fidelity inputs by $\tilde{\xi}_i^\hf\,{=}\,(\xi_i^\hf, 1)$ and $\tilde{\xi}_i^\lf\,{=}\,(\xi_i^\lf, 2)$, respectively.
Assuming conditional independence of all observations given the latent function values $\mathbf{\tilde{g}} = [(\mathbf{\tilde{g}}^\hf)^\top, (\mathbf{\tilde{g}}^\lf)^\top]^\top$, where
$\mathbf{\tilde{g}}^\hf=[ \tilde{g}(\tilde{\xi}_1^\hf),\dots,\tilde{g}(\tilde{\xi}_{n_\mathrm{\Xi,hf}}^\hf) ]^\top$ 
and
$\mathbf{\tilde{g}}^\lf=[ \tilde{g}(\tilde{\xi}_1^\lf),\dots,\tilde{g}(\tilde{\xi}_{n_\mathrm{\Xi,lf}}^\lf) ]^\top$, the joint likelihood factorizes as
\begin{multline} \label{eq:icm_joint_likelihood}
    p(\mathcal{D} \mid \mathbf{g})
    =
    \prod_{(i,j) \in \mathcal{C}^\hf} 
    \Phi\left(\frac{\tilde{g}(\tilde{\xi}^\hf_i) - \tilde{g}(\tilde{\xi}^\hf_j)}{\sqrt{2\sigma_\mathrm{n}^2}}\right)
    \\
    \cdot\prod_{(\xi_i^\lf, y_i^\lf) \in \mathcal{D}^\lf} \mathcal{N}\left(y_i^\lf ; \; \tilde{g}(\tilde{\xi}^\lf_i), \sigma_\mathrm{n}^2\right), 
\end{multline}
where preference observations are incorporated through the probit likelihood \eqref{eq:probit} and numerical observations through a Gaussian likelihood.
As the joint likelihood is non-Gaussian, the posterior must be approximated, as in the single-fidelity preference GP case (Section \ref{sec:fundamentals_gp_pref}).
Specifically, we employ MCMC-based approximations.

For the two-fidelity problem, we adopt the coregionalization matrix
\begin{equation} \label{eq:mm_icm_matrix_B}
    B = \begin{bmatrix}
        \sigma_\hf^2 
        & \rho\,\sigma_\hf\sigma_\lf
        \\
        \rho\,\sigma_\hf\sigma_\lf
        &
        \sigma_\lf^2 
    \end{bmatrix},
\end{equation}
where $\rho \in [0,1)$ and $\sigma_\hf, \sigma_\lf \in \mathbb{R}^+$ are hyperparameters.
This ensures that $B$ is positive definite.

\noindent We place a $\operatorname{Beta}(\alpha, \beta)$ prior on $\rho$ with $\alpha > \beta$, biasing the distribution toward values near $1$ to promote information transfer from low to high fidelity when only low-fidelity data have been collected.

\subsection{Multi-Modal AR1 GP Surrogates} \label{sec:ar1_mm}

In contrast to the intrinsic coregionalization approach, the AR1 scheme (cf. Section \ref{sec:fundamentals_ar1}) incorporates data of different modalities hierarchically through a two-step process. 
We first condition a standard GP \eqref{eq:gp_prior} on the numerical low-fidelity observations in the dataset $\mathcal{D}^\lf$ to obtain the low-fidelity model $g^\lf(\xi)$.
Next, we use the low-fidelity GP model to predict the mean and variance at the inputs $\Xi^\hf$, where high-fidelity comparative evaluations were performed.
For sampling-based inference via MCMC methods, these predicted statistics are computed empirically.

The high-fidelity model, which integrates both numerical and preferential data, is then expressed as
\begin{equation}
    g^\hf(\xi) = g^\lf(\xi) + \delta (\xi),
\end{equation}
where $\delta$ is a preference-based GP, and, in contrast to \eqref{eq:ar1_gp}, we omit the scaling parameter $\rho$.
This omission is justified because the scale of $g^\hf$ is negligible when conditioned only on comparative data, making $\rho$ redundant with the scaling parameter of the $\delta$-GP kernel.

It remains to infer the hyperparameters of the $\delta$-GP, as well as the vector  of latent values $\bm{\delta}$, where $[\bm{\delta}]_i = \delta(\xi^\hf_i),\, i=1,\dots,n_\Xi^\hf$ using MCMC sampling from the joint posterior
\begin{equation}
    p(\bm{\delta}, \theta  \mid  \mathcal{D}^\hf)  \propto
    p(\mathcal{D}^\hf  \mid  \bm{\delta}, \theta)
    \;
    p(\bm{\delta}  \mid  \theta)
    \;
    p(\theta),
\end{equation}
where $p(\theta)$ is the prior over hyperparameters, $p(\bm{\delta}  \mid  \theta)$ is the zero-mean GP prior, and $p(\mathcal{D}^\hf  \mid  \bm{\delta}, \theta)$ is the likelihood of the observed preference relations in $\mathcal{D}^\hf$. 
The likelihood for each comparison $c=(i,j) \in C^\hf$ is computed as 
\begin{multline}\label{eq:mmar1_pointwise_likelihood}
    p(c  \mid  \delta(\xi^\hf_i) , \delta(\xi^\hf_j)) = 
    \\
    \iiint \mathbf{1}_{\{g^\lf(\xi^\hf_i)+\delta(\xi^\hf_i)+\varepsilon_i \ge g^\lf(\xi^\hf_j)+\delta(\xi^\hf_j)+\varepsilon_j\}}(\varepsilon_i,\varepsilon_j,\mathbf{g}^\lf) \\ p\big(g^\lf(\xi^\hf_i), g^\lf(\xi^\hf_j)\big) \, p(\varepsilon_i) \, p(\varepsilon_j) \, \der\varepsilon_i \, \der\varepsilon_j \, \der \mathbf{g}^\lf,
\end{multline}
where $\varepsilon_i, \varepsilon_j \sim \mathcal{N}(0, \sigma_\mathrm{n}^2)$ represent independent additive noise accounting for inconsistencies in human decisions, and $\sigma_\mathrm{n}$ is a hyperparameter of the model.
By introducing $\eta = g^\lf(\xi^\hf_j) - g^\lf(\xi^\hf_i) + \varepsilon_j - \varepsilon_i$ with $\eta \sim \mathcal{N}( \Exp{g^\lf(\xi^\hf_j)} - \Exp{g^\lf(\xi^\hf_i)}, \Cov{g^\lf(\xi^\hf_j)-g^\lf(\xi^\hf_i)} + 2\sigma_\mathrm{n}^2)$, \eqref{eq:mmar1_pointwise_likelihood} simplifies to
\begin{multline}
        p(c  \mid  \delta(\xi^\hf_i) , \delta(\xi^\hf_j)) 
        = 
        \int \mathbf{1}_{\{\delta(\xi^\hf_i) - \delta(\xi^\hf_j) \ge \eta \}}(\eta) p(\eta) \, \der \eta 
        \\
        = \Phi\left(\frac{\delta(\xi^\hf_i) + \Exp{g^\lf(\xi^\hf_i)} - \delta(\xi^\hf_j) - \Exp{g^\lf(\xi^\hf_j)}}{\sqrt{\Cov{g^\lf(\xi^\hf_j)-g^\lf(\xi^\hf_i)} + 2\sigma_\mathrm{n}^2)}}\right),
\end{multline}
where $\Phi$, as previously, denotes the cumulative distribution function of the standard normal distribution.

Finally, we compute the posterior distribution under the multi-modal AR1 model using MCMC methods, as described in Section \ref{sec:fundamentals_gp_pref}.

\section{Example: Personalized Automated Driving}
\label{sec:example}

While automated driving systems can be designed mainly based on clearly defined, numerically tractable objectives --- such as minimizing travel time, reducing energy consumption, or maintaining desired safe distances from other vehicles --- they often must also account for more subjective goals, including passenger comfort \cite{steinke2025_diss}, perceived safety, or driving style.

In this work, we focus on the task of tuning an automated driving system according to individual passenger preferences, building on our previous study \cite{theiner2025_preflearnAD}.
The system employs an optimization-based trajectory planner followed by a trajectory tracking controller.
At each planning stage $k$, the planned trajectory $\hat{\tau}^*_{k}$ is obtained as the solution of an optimal control problem.
For details regarding the vehicle model and the formulation of the optimal control problem, we refer the reader to \cite{theiner2025_preflearnAD}.

To adapt the driving style to individual preferences, we learn parameters of the trajectory planner --- specifically, the weights of the cost function of the underlying optimal control problem --- using multi-modal multi-fidelity BO, as illustrated in Figure \ref{fig:overview_general}.
In particular, we optimize the weights associated with positive and negative longitudinal acceleration, lateral acceleration, longitudinal jerk, and lateral jerk, resulting in $n_\xi=5$ parameters.

\begin{figure}
    \centering
    \vspace{1mm}
    \includegraphics{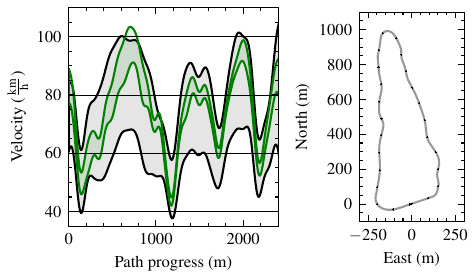}
    \caption{Left: Black and green areas show the $2\sigma$ credible intervals of the heteroscedastic Gaussian processes modeling observed human driver trajectories. The black model, trained on a range of drivers, serves as a low-fidelity model, while the green model, trained on one driver, is utilized to simulate high-fidelity preferences of a passenger for method evaluation. In each comparison, the simulated passenger prefers the trajectory that better aligns with the green model. Right: Overview of the track layout.}
    \label{fig:drivermodel_and_track}
\end{figure}

To facilitate sample-efficient learning, we construct a low-fidelity model of the human decision-making process from observed trajectories of human drivers.
Specifically, we proceed as in our previous work \cite{theiner2025_preflearnAD}, and model the range of observed velocity profiles using a heteroscedastic GP \cite{kesting2007_mostlikeliheteroskedasticgp}, see
the black shaded area in Figure \ref{fig:drivermodel_and_track}.
The resulting low-fidelity model of the human decision-making process is then expressed as the likelihood of a simulated trajectory under this probabilistic model.
This low-fidelity model provides a numerical approximation of human preferences, which can be combined with high-fidelity preferential feedback in the proposed multi-modal multi-fidelity BO framework.

In the following, we evaluate our method using a simulated passenger, where the true objective function underlying their preferences is known.
This is necessary for a quantitative evaluation using regret as a performance measure. Regret plots cannot be generated in realistic settings, where the true objective guiding the preferences is unknown.
High-fidelity preferential feedback is modeled using a heteroscedastic GP trained on a distinct set of trajectories from a single driver, as indicated by the green shaded area in Figure \ref{fig:drivermodel_and_track}.

To evaluate our proposed multi-modal multi-fidelity surrogates, we compare (i) standard preferential BO using a preferential GP surrogate model (cf. Section \ref{sec:fundamentals_gp_pref}), (ii) multi-modal multi-fidelity BO based on the proposed multi-modal ICM surrogate (cf. Section \ref{sec:icm_mm}), and (iii) multi-modal multi-fidelity BO based on the proposed multi-modal AR1 surrogate (cf. Section \ref{sec:ar1_mm}).
For the multi-modal methods, the optimization is split into a low-fidelity phase, where numerical observations of the low-fidelity objective are collected, and a high-fidelity phase, where the high-fidelity objective is optimized based on preferential feedback. 

All three procedures use the \emph{expected utility of the best option} (EUBO) acquisition function \cite{astudillo2023_qEUBO} to propose new comparison candidates during the high-fidelity phase.
The multi-fidelity methods additionally employ the \emph{integral predictive variance} (IPV) acquisition function \cite{seo2000_ipv} during the first 80 episodes of the low-fidelity phase, followed by 20 episodes using \emph{expected improvement} (EI) \cite{garnett2023_bo}.
This choice is consistent with the strategy outlined in Section \ref{sec:mfbo}, which involves mainly using exploration-intensive acquisition functions during the low-fidelity phase.

For all methods, posterior samples are drawn using the Pyro \cite{pyro-toolbox} implementation of the NUTS algorithm \cite{homan2014_mcmc_nuts}, an extension of the \emph{Hamiltonian Monte Carlo} (HMC) MCMC algorithm.

\begin{figure}
    \centering
    \vspace{1mm}
    \includegraphics{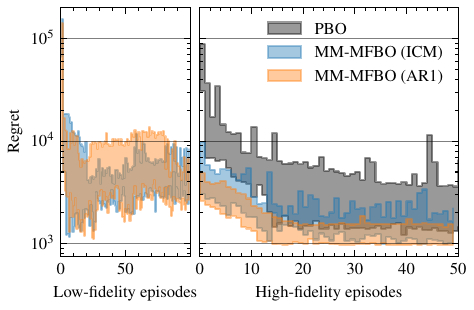}
    \vspace{-2mm}
    \caption{Regret of the recommended parameters $\xi^*_n$ --- obtained by maximizing the posterior mean of the surrogate model --- after each episode of the BO procedure. Shaded areas show the total range observed over five trials. Regret is computed using the true objective function $G(\xi)$ underlying the preferences of the simulated passenger.}
    \vspace{-2mm}
    \label{fig:regret_multi}
\end{figure}

Figure~\ref{fig:regret_multi} shows the regret of the three procedures over the course of the optimization runs. 
Regret is computed as the difference in performance between the trajectory induced by the best parameter recommendation after each episode --- obtained by maximizing the posterior mean of the surrogate --- and the best possible trajectory under the known objective.
Both proposed multi-fidelity methods clearly outperform the single-fidelity preferential Bayesian optimization, benefiting from successful information transfer from the preceding low-fidelity evaluations. 

In fact, the single-fidelity baseline requires approximately 10 to 15 high-fidelity episodes to reach regret scores that the proposed multi-modal multi-fidelity models achieve solely with low-fidelity data, before observing a single high-fidelity preference. 
Since each high-fidelity episode corresponds to a pair of closed-loop experiments, this corresponds to 20 to 30 closed-loop experiments.

In our evaluation, the autoregressive (AR1) structure achieves more consistent results and outperforms the ICM. 
We conjecture that this is due to the simpler and hierarchical correlation structure of the AR1 model, which naturally aligns with our setting.
In contrast, the correlation between fidelities in the ICM is more sensitive to variability in hyperparameter inference, which may lead to inconsistent performance across episodes.
This arises from the fact that the ICM does not a priori assume a hierarchical relation between the data sources, but treats them on equal footing.
The correlation structure must therefore be inferred from the data via hyperparameter estimation of the coregionalization matrix.
While this provides greater flexibility in modeling complex correlations compared to the AR1 model, accurately identifying the correlation structure from limited data can be challenging.

\begin{figure}
    \centering
    \vspace{1mm}
    \includegraphics{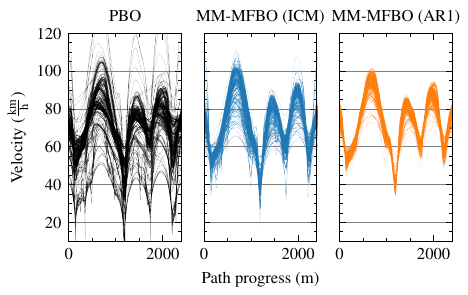}
    \caption{Trajectories sampled during the high-fidelity phases of each method, aggregated across all five experimental trials.}
    \vspace{-1mm}
    \label{fig:traj_samples}
\end{figure}

Finally, a similar trend can be observed when comparing the parameters queried by each method in every episode, rather than just the regret of the best recommendation over the episodes.
We emphasize that, in human-involved controller tuning, performance across all sampled comparisons is important: the human decision maker may experience discomfort when evaluating inferior parameterizations. 
Consequently, following our previous work \cite{theiner2025_preflearnAD}, we also examine the trajectories resulting from the parameters sampled across all high-fidelity episodes.
As shown in Figure \ref{fig:traj_samples}, the results indicate that the multi-fidelity structure significantly focuses the preference queries and reduces frequent sampling of inferior parameters. 

\pagebreak
\section{Conclusion}
\label{sec:conclusions}

Efficiently tuning controllers based on human preferences requires integrating additional information sources to minimize costly human feedback.
This work presents a framework for multi-modal, multi-fidelity Bayesian optimization that integrates numerical and preferential data to enable efficient learning of controller parameters.
Extending classical multi-fidelity Gaussian process formulations, we propose two surrogate model structures --- a coregionalization-based model and a hierarchical autoregressive AR1 model --- capable of jointly representing numerical evaluations and pairwise preference feedback.
Both surrogate structures facilitate information transfer across fidelities and modalities, thereby substantially improving data efficiency in settings where high-fidelity human feedback is limited.

The framework is demonstrated on a personalized automated driving task, where controller parameters are adapted to individual passenger preferences. 
The results indicate that both multi-modal, multi-fidelity surrogate models outperform standard preference-only Bayesian optimization by achieving faster and more consistent adaptation of the driving style. 
Among the two approaches, the AR1 structure exhibits particularly robust performance, suggesting that hierarchical formulations offer advantages when combining heterogeneous data modalities.

Finally, although demonstrated in automated driving, the proposed framework is general in nature and directly applicable to a wide range of controller tuning tasks involving human preferences, including exoskeletons, assistive robotics, autonomous driving, and other human-centered applications.

\bibliographystyle{IEEEtran}
\bibliography{references}

\end{document}